\documentclass{article}

\usepackage{PRIMEarxiv}

\usepackage[utf8]{inputenc} 
\usepackage[T1]{fontenc}    
\usepackage{hyperref}       
\usepackage{url}            
\usepackage{booktabs}       
\usepackage{amsfonts}       
\usepackage{nicefrac}       
\usepackage{microtype}      
\usepackage{lipsum}
\usepackage{fancyhdr}       
\usepackage{graphicx}       
\graphicspath{{media/}}     

\usepackage{multirow}
\usepackage{amsmath}
\usepackage{algorithm}
\usepackage{algpseudocode}

\title{Semi-Supervised Medical Image Segmentation with Co-Distribution Alignment
}

\author{
  Tao Wang $^{1,2,3}$, Zhongzheng Huang $^{2}$, Jiawei Wu $^{4}$, Yuanzheng Cai $^{1}$ and Zuoyong Li $^{1,}$\thanks{Corresponding author.
\textit{\underline{Citation}}: 
\textbf{Wang, T.; Huang, Z.; Wu, J.; Cai, Y.; Li Z. Semi-Supervised Medical Image Segmentation with Co-Distribution Alignment. Bioengineering 2023, 10(7), 869. https://doi.org/10.3390/bioengineering10070869}} \\
  $^{1}$ Fujian Provincial Key Laboratory of Information Processing and Intelligent Control, Minjiang University, China \\
  $^{2}$ College of Computer and Data Science, Fuzhou University, China \\
  $^{3}$ The Key Laboratory of Cognitive Computing and Intelligent Information Processing of Fujian Education Institutions,\\Wuyi University, China\\
  $^{4}$ School of Electrical and Mechanical Engineering, Fujian Agriculture and Forestry University, China\\
}

\begin{document}
\maketitle

\begin{abstract}
Medical image segmentation has made significant progress when a large amount of labeled data are available. However, annotating medical image segmentation datasets is expensive due to the requirement of professional skills. Additionally, classes are often unevenly distributed in medical images, which severely affects the classification performance on minority classes. To address these problems, this paper proposes Co-Distribution Alignment (Co-DA) for semi-supervised medical image segmentation. Specifically, Co-DA aligns marginal predictions on unlabeled data to marginal predictions on labeled data in a class-wise manner with two differently initialized models before using the pseudo-labels generated by one model to supervise the other. Besides, we design an over-expectation cross-entropy loss for filtering the unlabeled pixels to reduce noise in their pseudo-labels. Quantitative and qualitative experiments on three public datasets demonstrate that the proposed approach outperforms existing state-of-the-art semi-supervised medical image segmentation methods on both the 2D CaDIS dataset and the 3D LGE-MRI and ACDC datasets, achieving an mIoU of 0.8515 with only 24\% labeled data on CaDIS, and a Dice score of 0.8824 and 0.8773 with only 20\% data on LGE-MRI and ACDC, respectively.
\end{abstract}

\keywords{medical image segmentation \and semi-supervised learning \and distribution alignment \and co-training}

\section{Introduction}\label{sec:introduction}
Currently, there are various new technologies and devices that assist in clinical diagnostic work~\cite{bioengineering8110178, bioengineering8120199, bioengineering7030087}, among which medical image segmentation plays an important role in clinical auxiliary diagnosis~\cite{sym14101977}. Recently, researchers have made great efforts in medical image segmentation \cite{re:Unet, re:unet++,re:unetr} and achieved excellent performance with a large amount of labeled data. However, the annotation of medical data are typically dependent on medical professionals, and annotating large datasets {is time-consuming}.

To address this problem, semi-supervised (SS) medical image segmentation leverages a large amount of unlabeled data in conjunction with a small amount of labeled data to improve model performance. Particularly, unlabeled data are relatively affordable, as the laborious annotation process can be avoided. Recently, consistency regularization methods \cite{re:UAMT, re:CR2, re:misCT, re:dual-T2, re:loopUN} have received great attention in SS medical image segmentation. The primary difference of various consistency regularization methods lies in their intended objectives. For example, the majority of perturbation consistency methods \cite{re:misCT} tend to maintain a consistent unlabeled prediction with various augmentations. Uncertainty--aware methods \cite{re:UAMT, re:loopUN} force consistent predictions in reliable regions of two corresponding models. Multi-task based methods \cite{re:CR2, re:dual-T2} design a multi-task framework to guarantee an invariant relation of unlabeled data among different tasks. Despite the fact that these consistency regularization methods have obtained encouraging results, most of the previous methods neglect two essential problems: class imbalance and the mismatch of class distributions between labeled and unlabeled data. For the first problem, there have been some effective and efficient solutions~\cite{re:focal_loss,re:LT1} to deal with class imbalance in the fully supervised scenario. However, these methods are unsuitable for the semi-supervised case because long-tail samples and noisy samples are usually difficult to identify in unlabeled data. For the second problem, ReMixMatch~\cite{re:remixmatch} proposed a coefficient transform to align labeled and unlabeled class distributions for SS classification.
One of their key limitations, however, is that it only considers the empirical ground-truth class distribution, which could be highly imbalanced or even biased when the labeled data are scarce.
In addition, estimating the labeled distributions may lead to an unaffordable computational cost for dense prediction tasks such as image segmentation.

\begin{figure}[h!]
\centering
\includegraphics[width=0.7\linewidth]{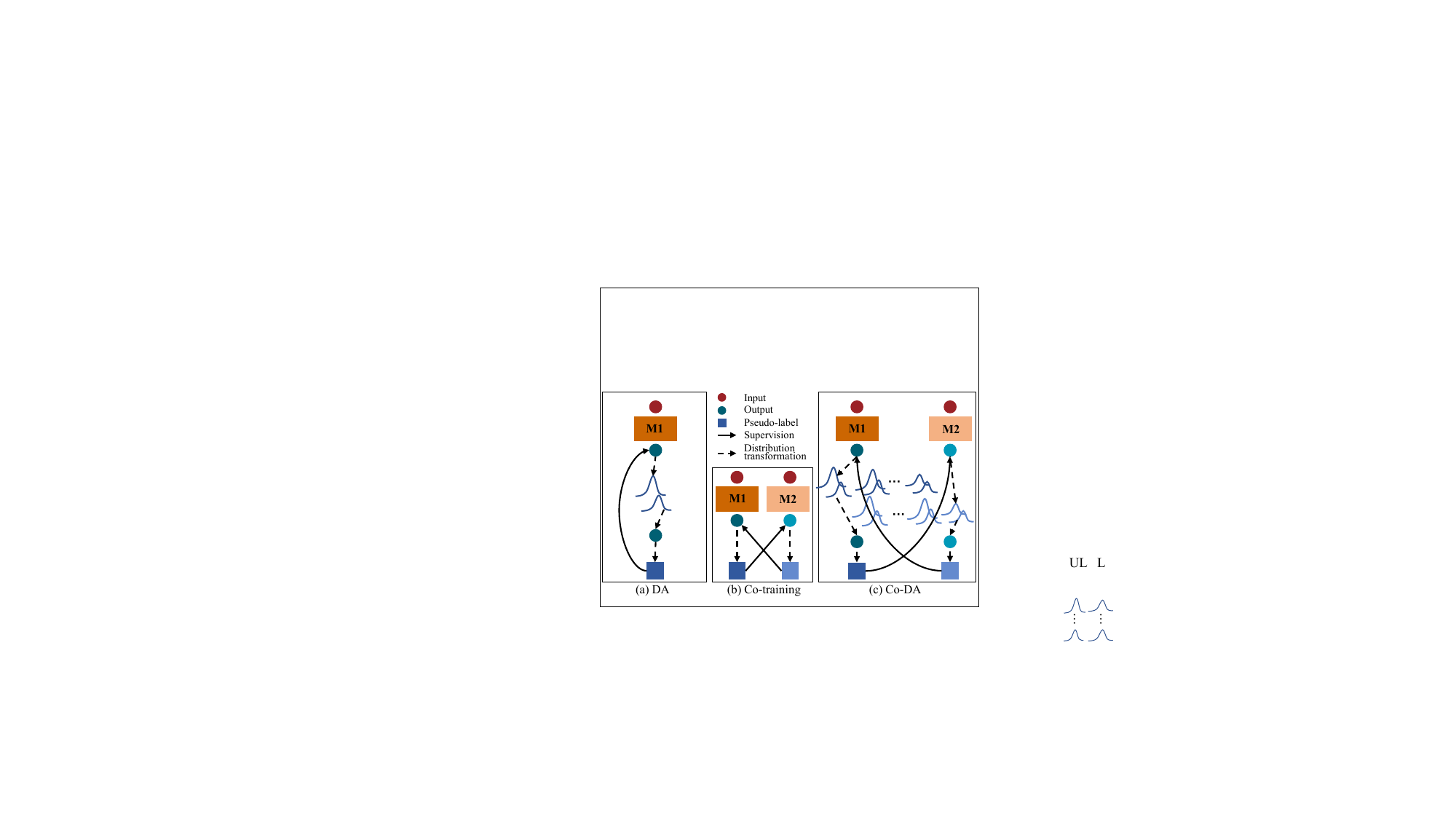}
\caption{Illustration of a comparison of Co-Distribution Alignment (Co-DA) with Distribution Alignment (DA) and co-training. \textbf{(a)} is the DA in ReMixMatch \cite{re:remixmatch}, \textbf{(b)} is co-training~\cite{re:CPS}, \textbf{(c)} is the proposed Co-DA. M1 and M2 here stand for two differently initialized networks.
The dark blue circles and squares denote the output and the pseudo-labels of M1. The light blue circles and squares denote the output and the pseudo-labels of M2.}
\label{fig:intro}
\end{figure}

In this paper, we focus on maintaining consistent distributions of labeled and unlabeled data under a co-training framework, unlike existing consistency regularization methods. However, it is non-trivial to achieve class distribution consistency. In particular, the empirical class distribution could be highly imbalanced or even biased when the data are sparsely labeled. Figure~\ref{fig:intro}a shows the vanilla Distribution Alignment (DA)~\cite{re:remixmatch} in which an overall output class distribution on unlabeled data are maintained to align model outputs to the empirical ground-truth class distribution. More specifically, DA maintains a running average of predictions on unlabeled data. When the model outputs a prediction for an unlabeled sample, the distribution alignment scales the prediction by the empirical class distribution on labeled data over the average predictions on unlabeled data, so as to obtain an output that is aligned to the ground-truth class distribution. On the other hand, Figure~\ref{fig:intro}b shows the cross-pseudo supervision~\cite{re:CPS} based on co-training~\cite{re:co-training} that uses two parallel networks with identical architecture but different initializations, and then uses the output of one network to supervise the other one.
Inspired by ReMixMatch and cross-pseudo supervision, we propose a novel Co-Distribution Alignment (Co-DA) method to overcome class imbalance and the mismatch of class distributions by integrating the above methods into a unified learning framework. Specifically, Co-DA transforms the model output according to the ratio of the class-specific marginal distribution on labeled data over the average model predictions on unlabeled data for that same class and supervise the model from the other view as pseudo-labels. Different from ReMixMatch, Co-DA uses an exponential moving average (EMA) to simplify the estimation of the labeled distribution.
More importantly, instead of relying solely on a single empirical ground-truth class distribution, we seek to fully exploit the model prediction for all classes, aiming to minimize the class-dependent distribution discrepancy between the model outputs on the labeled and the unlabeled data.
Therefore, we preserve independent distributions for each class instead of an overall distribution as shown in Figure \ref{fig:intro}a,c. On the other hand, in contrast to typical co-training, Co-DA tends to keep the consistency between labeled and unlabeled distributions rather than the cross-consistency of predictions as shown in Figure \ref{fig:intro}b,c. Therefore, Co-DA is more computationally efficient for dense prediction tasks such as image segmentation and is less likely to be affected by errors at individual pixels within the co-training framework. To further reduce the impact from inaccuracies in the model prediction on unlabeled data, we design an over-expectation cross-entropy loss to filter out noises in pseudo-labels.

Our main contributions can be summarized as follows:

\begin{itemize}
    \item To our knowledge, we are the first to solve the semi-supervised medical image segmentation problem with distribution alignment. In particular, we propose class-wise distribution alignment that utilizes the class-dependent output distribution instead of the overall empirical ground-truth class distribution, which could be highly imbalanced and biased when the labeled data are scarce.
    \item Our co-distribution alignment framework is more computationally efficient for dense prediction tasks involving a large number of pixels as compared to typical co-training methods such as CPS~\cite{re:CPS}. More importantly, distribution alignment has better regularization, as the proposed method provides superior performance.
    \item To further reduce the impact from inaccurate predictions on unlabeled data, we propose a simple, yet effective over-expectation cross-entropy loss to filter out noises in pseudo-labels.
    \item Experimental evaluation results on three publicly available medical imaging datasets demonstrate the superior performance of our approach compared to the state-of-the-art methods. Moreover, ablation studies also verify the efficacy of the various components in Co-DA.
    \item Our method does not depend on a particular deep network architecture. Therefore, it can be used in conjunction with different models for medical image segmentation as a plug-and-play module to address the challenges of learning from imbalanced data and the distribution mismatch between labeled and unlabeled data.
\end{itemize}

The rest of the paper is organized as follows. Section~\ref{sec:related} reviews recent literature in the areas of deep semi-supervised learning, semi-supervised medical image segmentation, co-training and distribution alignment methods. Section~\ref{sec:approach} describes the proposed method in detail, followed by experimental evaluation in Section~\ref{sec:experiments} and closing remarks in Section~\ref{sec:conclusion}.

\section{Related Work}
\label{sec:related}

In this section, we review related literature in semi-supervised medical image segmentation. We first review recent work in deep semi-supervised learning, and then more specifically in semi-supervised medical image segmentation, as well as progress in co-training and distribution alignment methods that are closely related to our approach.

\subsection{Deep Semi-Supervised Learning}
Recently, semi-supervised learning (SSL) has made remarkable progress in various machine learning tasks. SSL methods can be broadly categorized into pseudo-labeling methods, consistency regularization methods, entropy minimization methods and hybrid methods. Specifically, pseudo-labeling methods \cite{re:pseudo1, re:pseudo2, re:pseudo3, re:pseudo4} aim at obtaining the pseudo-labels of unlabeled data by self-training. Consistency regularization methods \cite{re:CR1, re:CR2, re:CR3, re:CR4} force similar predictions under different perturbations of unlabeled data to expand the decision regions. Entropy minimization methods \cite{re:entropyM1,re:entropyM2} tend to make the decision boundary follow low density regions with the help of unlabeled data. Hybrid methods \cite{re:fixmatch, re:remixmatch, re:meanteacher} simultaneously combine some advantages of the above SSL methods. Our method belongs to both pseudo-labeling and consistency regularization methods, yet it addresses a critical problem that is largely ignored in existing methods. To be specific, most existing methods neglect class imbalance and the mismatch of class distributions between labeled data and unlabeled data, which are common in medical images and severely affect the performance of SSL methods.

\noindent\subsection{Semi-Supervised Medical Image Segmentation}
The distinctive appearance and class distribution characteristics of medical images pose unique challenges in applying SSL methods to them. In particular, medical image segmentation usually involves localizing objects with extreme shape and scale variations, and the class distribution could be highly skewed. Similar to generic SSL methods, common semi-supervised medical image segmentation methods include GAN-based, consistency regularization and pseudo-labeling methods. Specifically, GAN-based methods~\cite{re:gan1, re:gan2} attempt to use adversarial training to fool the discriminator with unlabeled data. For example, DCT-Seg \cite{re:DCOT} utilizes two models that are co-trained to generate pseudo-labels for each other. In addition, SS-Net~\cite{re:CLSS} proposes a collaborative learning method to jointly improve the performance of disease grading and lesion segmentation with an attention mechanism. On the other hand, different from consistency regularization methods in generic SSL, for medical image segmentation, people usually design a strategy to keep consistency in local regions, e.g., regions of low uncertainty~\cite{re:loopUN}, regions of random category~\cite{re:CRcl}, etc. Finally, pseudo-labeling methods leverage an auxiliary model to generate pseudo-labels for unlabeled data. Unlike existing methods, our Co-DA focuses on exploring the discrepancy between labeled and unlabeled data distributions, which is of vital importance in SS medical image segmentation due to data scarcity and imbalance.

\noindent\subsection{Co-Training}
The original co-training algorithm~\cite{re:co-training} assumes that there are two naturally segmented views of the same instance, which are redundant and independent under certain conditions. Concretely, data from any of them are sufficient to train a strong learner, and the views are independent of each other. More specifically, the main steps of the co-training algorithms are divided into view acquisition, learner differentiation and label confidence estimation~\cite{re:reviewCo}. Recent progresses in co-training~\cite{re:semiCO1, re:DCOT,re:CO3} primarily focus on maintaining the diversity across models with deep networks. In particular, cross-pseudo supervision~\cite{re:CPS} proposes to impose consistency on two segmentation networks perturbed with different initialization for the same input image. In contrast, Co-DA uses co-training to align the distributions of labeled and unlabeled predictions.

\noindent\subsection{Distribution Alignment}
Distribution alignment is widely used in domain adaptation. These methods work by aligning marginal distributions~\cite{re:MDA1, re:MDA2} or joint distributions~\cite{re:JDA1, re:JDA2, re:JDA3} of different domains. In semi-supervised learning, domain alignment has also been considered to close the gap between predictions on labeled and unlabeled data. Compared to our proposed approach, the most relevant existing method is ReMixMatch~\cite{re:remixmatch}, which uses a coefficient transformation to align the marginal distributions of labeled and unlabeled data. However, class distribution in medical images can be highly imbalanced, while ReMixMatch treats samples from all classes as a whole and neglects the variations in marginal distribution of each class. Different from ReMixMatch, in this work, we align marginal distributions in a class-wise manner that is more general and works better for imbalanced datasets with minority classes.

\section{Our Approach}
\label{sec:approach}

\begin{figure*}[ht]
\centering
\includegraphics[width=0.8\linewidth]{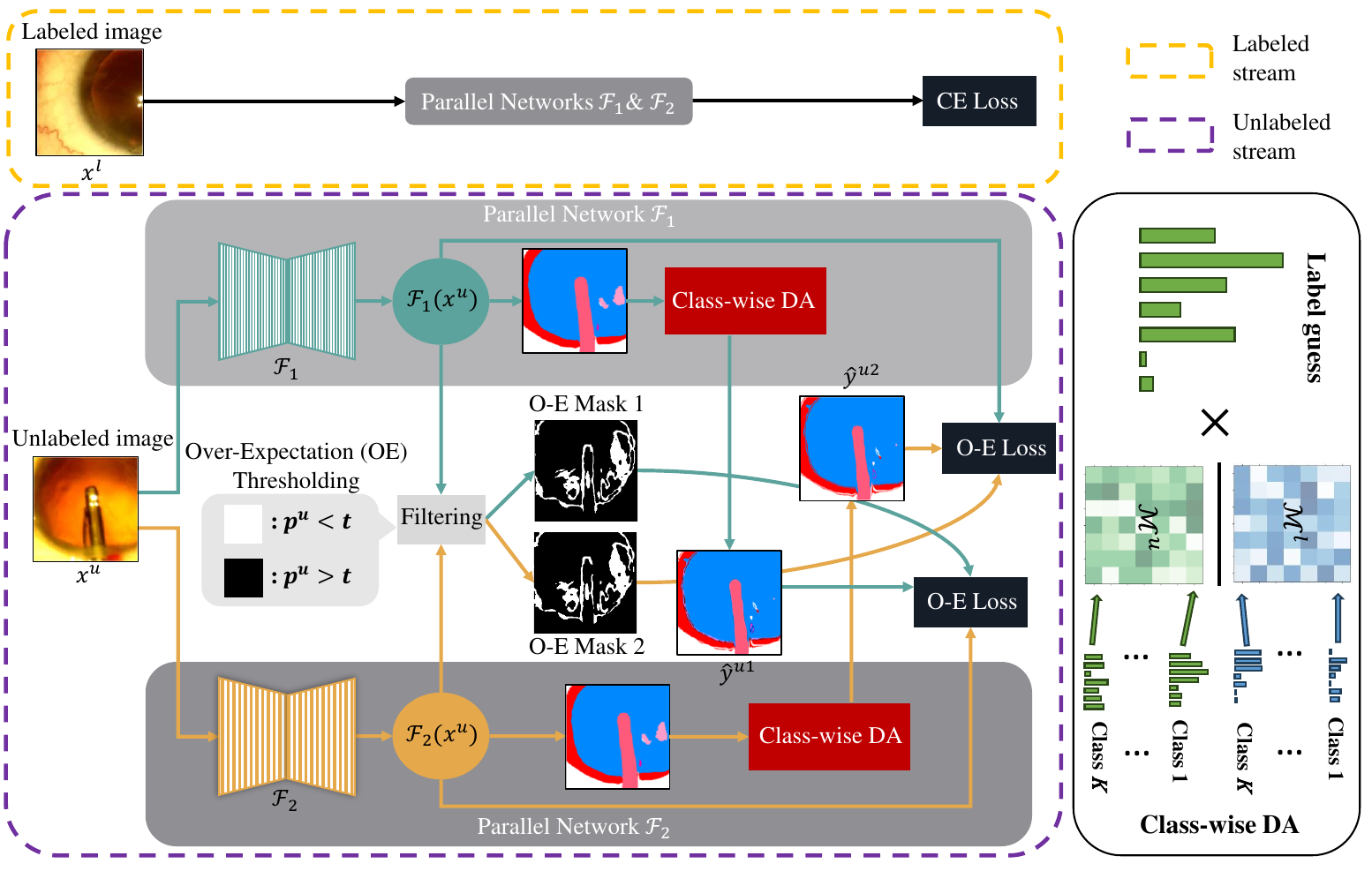}
\caption{Overview of Co-DA for semi-supervised medical image segmentation, which consists of a stream for labeled data and a stream for unlabeled data. $\mathcal F_1$ and $\mathcal F_2$ are two parallel networks with the same architecture but different initializations. For labeled data, the standard cross-entropy loss (CE Loss) is used. For unlabeled data, $\hat{y}^{u1}$ and $\hat{y}^{u2}$ are pseudo-labels given by Equation~\eqref{eq:T_D}, and we use the aligned output distribution from one network to supervise the other one. For the over-expectation cross-entropy loss (O-E Loss) in Equation~\eqref{eq:un_loss}, $p^u=\max(\mathcal F(x^u))$ denotes the class probability corresponding to the pseudo-labels, $t$ stands for the threshold given by Equation~\eqref{condition_threshold} and the O-E Masks are the thresholded binary masks to filter out pixels with low confidence in their pseudo-labels. We use class-wise DA to align marginal distributions for each class to better deal with imbalanced datasets. Here, $\mathcal M^l$ and $\mathcal M^u$ denote the labeled and the unlabeled marginal distributions.}
\label{fig:pathdemo2}
\end{figure*}

In this section, we describe the proposed approach in detail. We begin by revisiting the cross-pseudo supervision framework that our work is based on. Afterwards, we describe the three main components of Co-Distribution Alignment, i.e., marginal distribution estimation, class-wise distribution estimation and distribution transformation. In addition, we introduce an over-expectation cross-entropy loss that is used to further improve model performance by filtering out inaccurate pseudo-labels. The overall framework is presented in Figure~\ref{fig:pathdemo2}.

\subsection{Cross-Pseudo Supervision}
\label{sec:cps}
Cross-pseudo supervision works by generating two views of the same input image with differently initialized segmentation networks. In this way, one network can discover the self-mistakes made by the other model~\cite{re:co-training, re:CPS}. Following~\cite{re:CPS}, we also use two parallel networks~$\mathcal F_1$, $\mathcal F_2$ with the same structure but different initialization for collaborative training. To define our problem more formally, and without loss of generality, we view the medical image segmentation problem as a pixel-wise classification one since our focus in this paper is the cross-alignment of marginal distributions. More specifically, let $\mathcal{D}_l=\{(x_i^l, y_i^l)\}_{i=1}^m$ denote the labeled dataset and $\mathcal{D}_u=\{x_j^u\}_{j=1}^n$ the unlabeled dataset. Here, $x_i^l$ and $x_j^u$ are the image pixels, and $y_i^l$ is the class label for $x_i^l$.
The loss function on the labeled dataset can be written as:
\begin{equation}
    \mathcal L_s= -\frac{1}{m} \sum_{i=1}^m y_i^l \left[\log \mathcal F_1(x_i^l) + \log \mathcal F_2(x_i^l)\right],
    \label{eq:labelLoss}
\end{equation}

For the unlabeled data, we adopt the cross-pseudo supervision loss, which uses the pseudo-labels generated by one model to supervise the prediction of the other model, as follows:
\begin{equation}
    \label{eq:crossUn}
    \mathcal L_u= -\frac{1}{n} \sum_{j=1}^n \left [ \hat{y}_j^{u2} \log \mathcal F_1(x_j^u) +\hat{y}_j^{u1} \log \mathcal F_2(x_j^u) \right ],
\end{equation}
where $\hat{y}_j^{u1}$, $\hat{y}_j^{u2}$ are the pseudo-labels generated by $\mathcal F_1$ and $\mathcal F_2$, respectively. Here, pseudo-labels are the mode of prediction, i.e., the label derived from the most probable class.

Equation~\eqref{eq:crossUn} shows that pseudo-labels determine our optimization objective. Therefore, most existing methods~\cite{re:CPS, re:CO3} focus on correcting or generating more accurate pseudo-labels to supervise unlabeled predictions. However, their methods neglect the fundamental challenges of learning from imbalanced data and the distribution mismatch between labeled and unlabeled data. In the next section, we introduce the proposed Co-DA that specifically addresses these problems.

Let us take a closer look at the pseudo-labels in Equation~\eqref{eq:crossUn}. We can conjecture that, ideally, if the pseudo-labels $\hat{y}_j^{u1}$, $\hat{y}_j^{u2}$ follow the class distribution on labeled data, applying Equation~\eqref{eq:crossUn} would already align the unlabeled class distribution to the labeled class distribution. However, due to class imbalance in medical image segmentation, the distribution of pseudo-labels could be biased toward majority classes. In this paper, we propose to address this problem by explicitly aligning the marginal distributions on unlabeled and labeled data after cross-pseudo supervision. In particular, our Co-DA considers $K$ class-dependent distributions (where $K$ is the number of classes) instead of a single overall empirical class distribution in naive DA. This approach allows us to capture and align finer class-wise statistics in the marginal distributions, as we describe in the next section.

\subsection{Co-Distribution Alignment}

In this section, we outline the proposed Co-Distribution Alignment in detail. Specifically, we first introduce an efficient method to estimate the class distribution on labeled data during training, and then discuss class-wise distribution estimation for capturing detailed statistics in labeled and unlabeled marginal distributions, followed by the distribution transformation to align the unlabeled distributions to labeled distributions.

\subsubsection{Marginal Distribution Estimation}
The vanilla DA in ReMixMatch~\cite{re:remixmatch} needs to use the empirical ground-truth class distribution for aligning model predictions on unlabeled data to it. However, estimating ground-truth class distribution becomes computationally expensive for segmentation tasks where a single image contains a large number of pixels, and using only a small amount of labeled data is inappropriate, as the estimation may not well represent the overall class distribution. To address this issue, we use the exponential moving average (EMA) to estimate the aggregated class predictions on labeled data, which can be written as:
\begin{equation}
    \label{eq:ema_label}
    \mathcal T^{l} = \alpha \times \hat{\mathcal T^l} + (1-\alpha) \times \mathbb{E}[\mathcal{F}(x^l;\Theta)],
\end{equation}
where $\mathcal{T}^l$ and $\hat{\mathcal{T}^l}$ are the labeled distributions of the current iteration and the previous iteration, $\alpha$ determines that $\mathcal{T}^l$ is the average prediction of last $1/(1-\alpha)$ iterations and $\Theta$ represents the parameters of the network. It should be noted that our work differs from ReMixMatch~\cite{re:remixmatch} in that we use EMA to estimate the class predictions on labeled data, which is more computationally efficient for dense prediction tasks. In addition, we note that the MLE of observations can be given by:
\begin{equation}
    \begin{split}
        \Theta &= arg\max_{\theta} \mathcal {F}(\mathcal{X};\theta) \\
        &= arg\min_{\theta} -\sum_{i=1}^m \log \mathcal {F}(x^l_i;\theta) - \sum_{i=1}^n \log \mathcal {F}(x_i^u;\theta) ,
    \end{split}
\end{equation}

\noindent where $\mathcal{X}$ is the entire dataset that includes both labeled and unlabeled data. Obviously, $\Theta$ is related to the likelihoods of both labeled and unlabeled data. Therefore, the labeled distribution estimated by Equation~\eqref{eq:ema_label} considers both labeled and unlabeled data to estimate the labeled distribution and only costs manageable computations during training.

\subsubsection{Class-Wise Distribution Estimation}
An important limitation of the original DA is that class imbalance may cause the tail classes to diminish from the estimated distribution, as they only represent a small fraction of labeled data, ultimately leading to a biased distribution transformation. Unlike naive DA~\cite{re:remixmatch}, Co-DA builds $K$ independent distributions for each class, where $K$ is the number of classes and each distribution is class-specific. Therefore, Co-DA is more robust to class imbalance due to the decoupled modeling process. Specifically, we maintain two matrices $\mathcal M^l \in \mathbb{R}^{K \times K}$ and $\mathcal M^u\in \mathbb{R}^{K \times K}$ for labeled and unlabeled distributions, respectively, in which row $i$ is the distribution for the $i$-th class. For the $i$-th row of the labeled matrix $\mathcal M^l_i$, we use Equation~\eqref{eq:ema_label} to approximate the marginal distribution according to the ground-truth class label as follows:
\begin{equation}
    \mathcal M^l_i = \alpha \cdot \hat{ \mathcal M}^l_i + (1-\alpha) \cdot \mathbb{E}[\mathbb{I}_{[y^l=i]} \mathcal F(x^l)],
    \label{update_label}
\end{equation}

\noindent where $\mathbb{I}_{[\cdot]}(\cdot)$ is the indicator function. For the $i$-th row of the unlabeled matrix $\mathcal M^u_i$, we update the marginal distribution by EMA according to the model prediction following ReMixMatch \cite{re:remixmatch}. The class membership of unlabeled data are obtained by the mode of prediction, i.e., the most probable class. However, we observe that the tail categories in the unlabeled distributions may be difficult to update due to their infrequent presence in a batch. In this case, we use the inverse transformation from labeled distribution to unlabeled distribution to update the unlabeled distribution. The overall update strategy of $\mathcal M^u_i$ can be written as follows:
\begin{equation}
    \mathcal M^u_{i}=\begin{cases}
        \hat{\mathcal M^l_i} \times \mathbb{E}[\hat{\mathcal M^u_i} / \hat{\mathcal M^l_i}], & \text{if } \{\hat{y}^u | \hat{y}^u=i\} = \emptyset, \\ \\
        \small \alpha \times \hat{ \mathcal M}^u_i + (1-\alpha) \\ \qquad \times \mathbb{E}[\mathbb{I}_{[\hat{y}^u=i]} \mathcal F(x^u)], & \text{otherwise},\\
    \end{cases}\label{update_unlabel}
\end{equation}
\noindent where $\hat{y}^u = \arg\max \mathcal F(x^u)$ and $\hat{ \mathcal M}^u_i$ is the unlabeled distribution in the previous iteration. This update strategy stems from the original distribution alignment proposed in ReMixMatch \cite{re:remixmatch}, but here, we use it  to estimate the unlabeled distribution when the unlabeled data from a certain class (usually the minority classes) are absent in a batch. In addition, we use EMA to maintain a stable update of the distributions required for our Co-DA in the training process. It should be noted that, in practice, there are two models (i.e., $\mathcal F_1$ and $\mathcal F_2$) in our method, and each model will be used to update their own $\mathcal M^l$ and $\mathcal M^u$; we omit their subscripts for notational simplicity.

\subsubsection{Distribution Transformation}
Following ReMixMatch, we align the distributions using coefficient transformation but with each class functioning independently. Besides, we use a temperature $\tau$ to scale the labeled distribution following CReST \cite{wei2021crest}. Different from CReST, however, we choose an adaptive temperature according to the class-specific aggregated model prediction. More specifically, the temperature and the transformation of class $i$ is given by:

\begin{equation}
    \tau_{i} = 1 - \mathcal M^l_{ii}, \label{temperature}
\end{equation}
\begin{equation}
    [\hat{y}^u=i] \implies \mathcal F'(x^u) = \text{Normalize}\big( \mathcal F(x^u) \otimes {\mathcal M^l_i}^{\tau_i} \oslash \mathcal M^u_i \big)
    \label{eq:T_D}
\end{equation}

\noindent where $[\hat{y}^u=i]$ denotes that, according to the network, the $i$-th class is the most probable, and we therefore use the $i$-th row in $\mathcal M^l$ and $\mathcal M^u$, i.e., $\mathcal M^l_i$ and $\mathcal M^u_i$, for the coefficient transformation. In Equation~\eqref{eq:T_D}, $\otimes$ and $\oslash$ denote the element-wise product and division, respectively. In addition, $\mathcal{F}(x^u)$ and $\mathcal{F}'(x^u)$ denote the network prediction for $x^u$ before and after distribution transformation.

We note that the temperature $\tau_i$ further prevents the transformed distribution to be dominated by majority classes. Specifically, $\tau_i \rightarrow 0$ makes the labeled distribution closer to uniform distribution, while a larger $\tau_i$ results in a smaller shift. $\tau_i$ not only makes the model more robust against noises in pseudo-labels in the early training stage, but also encourages the emergence of minority classes in the middle and late stages.

After distribution transformation, the network prediction after alignment, $\mathcal F'(x^u)$, is then used for cross-pseudo supervision according to Equation~\eqref{eq:crossUn}, instead of $\mathcal F(x^u)$ as originally shown in Equation~\eqref{eq:crossUn}.

\subsection{Over-Expectation Cross-Entropy Loss}
In order to further reduce the negative impact of inaccurate pseudo-labels, we propose an over-expectation cross-entropy loss for learning from unlabeled data. Motivated by the definition of EMA, $\mathcal M^l$ and $\mathcal M^u$ can be regarded as the expectations of the labeled and unlabeled distributions, respectively. Intuitively, we can use the estimated aggregated model prediction for the $i$-th class as an adaptive threshold to filter out unlabeled samples below expectations to reduce noise. More specifically, the threshold $t(i)$ for the $i$-th class is given by:
\begin{equation}
    t(i)=\mathcal M^u_{ii}
    \label{condition_threshold}
\end{equation}

\noindent As we will show in Section~\ref{sec:abl}, this adaptive and dynamic threshold is consistently superior to different static threshold values and do not bring in additional hyperparameters. In this way, Equation~\eqref{eq:crossUn} can be rewritten as:
\begin{equation}
\begin{split}
    \mathcal L_u= -\mathbb{E} \big[\mathbb{I}_{[p^{u2}_i > t({\hat{y}^{u2}_i})]} \hat{y}_i^{u2} \log \mathcal F'_1(x_i^u)  + \mathbb{I}_{[p^{u1}_i > t({\hat{y}^{u1}_i})]} \hat{y}_i^{u1} \log \mathcal F'_2(x_i^u) \big],
    \label{eq:un_loss}
\end{split}    
\end{equation}

\noindent where $p^u_i$ stands for the probability corresponding to the pseudo-labels, $\hat{y}^u_i$ stands for the class prediction of the pseudo-labels and we refer to our loss in Equation~\eqref{eq:un_loss} as the over-expectation cross-entropy loss (O-E Loss). In a nutshell, this is an extension to the soft label cross-entropy loss that incorporates the threshold filtering in Equation~\eqref{condition_threshold}. The logarithm comes from the cross-entropy between two probability distributions, and we refer readers to~\cite{shore1981properties} for further background information. See Figure~\ref{fig:pathdemo2} for the O-E Masks 1 and 2 as an illustration for applying Equation~\eqref{eq:un_loss} in practice. The complete training process of Co-DA is summarized in Algorithm~\ref{alg:algorithm1}.

\begin{algorithm}[h]
\caption{Co-Distribution Alignment.}
\label{alg:algorithm1}
\begin{algorithmic}[1] 
\State \textbf{Input}: Labeled training data $\{(x^l_i,y^l_i)\}_{i=1}^m$, unlabeled training data $\{x_i^u\}_{i=1}^n$, max iterations $I_{max}$, two parallel networks $\mathcal F_1(\Theta_1)$ and $\mathcal F_2(\Theta_2)$, learning rate $\eta$, momentum $\alpha$ for exponential moving average
\State Initialize $\mathcal F_1(\Theta_1)$ and $\mathcal F_2(\Theta_2)$ with different parameters
\For{iteration=1 to $I_{max}$}
\State Update the labeled distribution $\mathcal M^l$ by Equation~\eqref{update_label};
\State Update the unlabeled distribution $\mathcal M^u$ by Equation~\eqref{update_unlabel};
\State Generate the pseudo-labels $\hat{y}^{u1}$ and $\hat{y}^{u2}$ w.r.t. $\mathcal F_1(\Theta_1)$ and $\mathcal F_2(\Theta_2)$ by Equation~\eqref{eq:T_D};
\State Calculate the supervised loss $\mathcal L_s(x^l,y^l)$ according to Equation~\eqref{eq:labelLoss};
\State Calculate the unsupervised loss $\mathcal L_u(x^u,\hat{y}^{u1},\hat{y}^{u2})$ according to Equation~\eqref{eq:un_loss};
\State Update $\Theta_1$= $\Theta_1$ - $\eta \cdot \nabla_{\Theta_1} (\mathcal L_s + \mathcal L_u)$;
\State Update $\Theta_2$= $\Theta_2$ - $\eta \cdot \nabla_{\Theta_2} (\mathcal L_s + \mathcal L_u)$;
\EndFor \\
\textbf{Output}: $\mathcal F_1(\Theta_1)$ and $\mathcal F_2(\Theta_2)$.
\end{algorithmic}
\end{algorithm}

\section{Experiments}
\label{sec:experiments}

In this section, we thoroughly verify the efficacy of the proposed Co-DA on three challenging public medical image segmentation datasets. Specifically, we first outline the experimental setup, including an introduction to the datasets, the evaluation metrics and our implementation details in Section~\ref{sec:setup}, followed by quantitative and qualitative results on the three datasets in Section~\ref{sec:res_cadis}, Section~\ref{sec:res_lge-mri} and Section~\ref{sec:res_acdc}, respectively. We also present ablation studies in Section~\ref{sec:abl} to demonstrate that our method provides a strong performance that is comparable to its fully supervised variant, and the individual components proposed in our method are all contributing to the performance of our model.

\subsection{Experimental Setup}

\label{sec:setup}
\subsubsection{CaDIS}
We first evaluate the performance of our method on a 2D medical image segmentation task. The publicly available CaDIS dataset~\cite{re:cadis-bouget2017vision, re:cadis-trikha2013journey} consists of 4671 frames from 25 surgical videos, which are collected by experts and annotated at the pixel level. Following~\cite{re:cadis-bouget2017vision}, we consider three progressively more difficult semantic segmentation tasks on this dataset. Specifically, task 1 contains 8 classes, with 4 for anatomical structures, 1 for all instruments and 3 for other objects that appear in frames; task 2 contains 17 classes, which divides the single instrument classes in task 1 into 10 more specific classes of instruments; task 3 contains 25 classes, where instruments are further subdivided according to the handles and parts of certain instruments.

To thoroughly demonstrate the efficacy of the proposed method, we compare our method with state-of-the-art competing methods on all three tasks with varying levels of labeled data. This task poses some unique challenges for segmenting the anatomical structures, surgical instruments and other objects (i.e., surgical tapes, hands and eye retractors) simultaneously. In particular, classes are unevenly distributed, and some objects are either thin or small, or both.

\subsubsection{Late Gadolinium Enhancement MRI}
To further demonstrate the performance of our method on 3D medical image segmentation tasks, we also evaluate our method on the {Late Gadolinium Enhancement MRI (LGE-MRI)} dataset~\cite{re:la-xiong2021global}. This dataset is a collection of 154 3D LGE-MRIs acquired from the Left Atrial Segmentation Challenge, which contains data from 60 patients with atrial fibrillation prior to and post-ablation treatment. The goal is to perform Left Atrium (LA) segmentation, and the images have an isotropic resolution of 0.625 $\times$ 0.625 $\times$ 0.625 {mm}$^3$. In particular, it is challenging to segment the LA in the top and the bottom slices, corresponding to the pulmonary veins and the mitral valve, respectively.

\subsubsection{ACDC}
In addition, we evaluate the performance of our method on another 3D medical image segmentation task using the ACDC (Automated Cardiac Diagnosis Challenge) 2017 dataset~\cite{bernard2018deep}. This dataset consists of MRI images from 100 patients with expert annotations. Among these, 2 are used as the validation set, 20 are used as the test set and the rest are used as the training set. The ACDC dataset contains images from patients with normal cardiac anatomy as well as those with previous myocardial infarction, dilated cardiomyopathy, hypertrophic cardiomyopathy and an abnormal right ventricle. One unique challenge in this task lies in identifying the left ventrice, the myocardium and the right ventrice, which could all be very small at certain anatomical structures such as the apex.

\subsubsection{Evaluation Metrics}
This section presents the evaluation metrics employed to assess the effectiveness of the proposed approach. Firstly, we follow~\cite{re:pissas2021effective} and use the mean Intersection over Union (mIoU) to evaluate the CaDIS dataset, given as follows:
 \begin{equation}
     \text{mIoU} = \frac{1}{c}{\sum\limits_{i = 0}^{c}\frac{\text{TP}}{\text{TP} + \text{FP} + \text{FN}}},
 \end{equation}

where $c$ is the number of classes, and TP, FP and FN denote the number of pixels that are true positives, false positives and false negatives, respectively. In addition,
Dice score, Jaccard, the average surface distance (ASD) and the 95\% Hausdorff Distance (95HD) are adopted to evaluate LGE-MRI~\cite{re:UAMT}. Also, methods are evaluated with Dice score and 95HD for ACDC 2017. Specifically, Dice, Jaccard and ASD are given as follows:
\begin{equation}
    \text{Dice} = {{2\text{TP}}/(2\text{TP} + \text{FP} + \text{FN})},
\end{equation}
\begin{equation}
    \text{Jaccard} = {{\text{TP}}/(\text{TP} + \text{FP} + \text{FN})},
\end{equation}
\begin{equation}
\begin{split}
    \text{ASD} = \frac{\left( {\sum\limits_{a \in S{(A)}}{\min\limits_{b \in S{(B)}}\left| \left| {a - b} \right| \right|}} + {\sum\limits_{b \in S{(B)}}{\min\limits_{a \in S{(A)}}\left| \left| {b - a} \right| \right|}} \right)}{\left| {S(A)} \right| + \left| {S(B)} \right|},
\end{split}
\end{equation}
where $S(\cdot)$ denotes the set of surface voxels, and the two sets $A$ and $B$ refer to the ground-truth and network prediction, respectively. $||\cdot||$ denotes the L2 distance. For the Hausdorff distance, it could be written as:
\begin{equation}
    \text{HD} = \max\left( h\left( {S(A),S(B)} \right),h\left( S(B),S(A) \right) \right),
\end{equation}
where $h\left( {S(A),S(B)} \right) = \underset{a \in S{(A)}}{\max}\left\{ {\underset{b \in S{(B)}}{\mathit{\min}}\left| \left| {a - b} \right| \right|} \right\}$, denoting the maximum distance of voxels in $A$ to the nearest voxel in $B$. The 95\% Hausdorff Distance is based on the 95th percentile of the surface distance above in order to eliminate the impact from a small number of outliers.

\subsubsection{Implementation Details}

For CaDIS, we use the publicly available split strategy \cite{re:pissas2021effective} of 3550 frames for training and the remaining 1120 for validation. For $12\%$, $24\%$ and $49\%$ labeled data in our experiments, we randomly choose 424, 834 and 1729 labeled frames, respectively. In addition, we randomly crop the original image to $352 \times 352$ and augment the data in the same way as in \cite{re:pissas2021effective}, including random horizontal flipping, 
Gaussian noise and color jittering. To mitigate the effects of data imbalance, we also follow \cite{re:pissas2021effective} and use Repeat Factor Sampling in the labeled set. 
For LGE-MRI, we are consistent with \cite{re:UAMT}'s pre-processing scheme, dividing the original 100 sheets of data into 80 for training and 20 for validation. All data are centrally cropped in the heart region and randomly cropped to $112 \times 112 \times 80$. The data augmentation strategy used included random flips and rotations of 90, 180 and 270 degrees. For the ACDC 2017 dataset, we use random flip, rotation and scaling as data 
augmentation.

The proposed method is implemented with PyTorch \cite{re:paszke2019pytorch}. We use an nVIDIA GeForce RTX 3090 GPU for training. We choose to train CaDIS and ACDC datasets with EfficientNet-B3-based \cite{re:tan2019efficientnet} Unet \cite{re:Unet} and LGE-MRI with Vnet \cite{re:vnet-milletari2016v} as backbone architectures. Both networks are updated using the SGD optimizer with a momentum of 0.9 and an initial learning rate of 0.0001, with the learning rate divided by $0.001^{1/epoch}$ for each epoch, where epoch depends on the ratio of iterations to the number of labeled data in training, and the batch size is set to 8. For CaDIS, we set the number of iterations in one epoch to 10,000 when the ratio of labeled data are 12\%, 20,000 when the ratio of labeled data are 24\% and 50,000 when the ratio of labeled data are 49\%; for training LGE-MRI, the number of iterations is set to 6000 uniformly; for ACDC, the number is 30,000 uniformly.

\subsection{Results on CaDIS}
\label{sec:res_cadis}
In order to better verify the effectiveness of the proposed method, we compare it with the following state-of-the-art methods: URPC \cite{re:urpc-luo2022semi}, UAMT \cite{re:UAMT}, CLD \cite{re:CLD-lin2022calibrating} and CPS~\cite{re:CPS}. Among these methods, URPC \cite{re:urpc-luo2022semi}, UAMT \cite{re:UAMT} and CLD \cite{re:CLD-lin2022calibrating} are specifically designed for semi-supervised medical image segmentation, while CPS \cite{re:CPS} is a generic semantic segmentation algorithm. We also experiment with a naive setup, i.e., to train the model with only labeled data, denoted as Baseline. As shown in Table \ref{table:cadis}, Co-DA outperforms other methods on different split settings. Specifically, in Task 1, our method is able to improve 0.1223, 0.1017 and 0.0329 on mean IoU with 12\%, 24\% and 49\% labeled data, respectively. For Task 2 and Task 3, the corresponding improvements are 0.1304, 0.1228, 0.1832 and 0.0845, 0.1181, 0.1189. It should be noted that under Task 1 with 12\% labeled data, the performance of our method is slightly lower than UAMT for semi-supervised medical image segmentation and CPS for generic image segmentation, which is probably due to the data distribution being relatively balanced when the amount of data and number of classes are small. However, as the problem gets more difficult when we have more imbalanced datasets with more classes, the performance gain obtained from our method becomes more prominent.

\begin{table*}[h!]
\caption{Mean IoU results on CaDIS. Bold numbers represent the best performance.}
\centering
\begin{tabular}{l|ccc|ccc|ccc}
\hline
\multicolumn{1}{c|}{\multirow{2}{*}{Method}} & \multicolumn{3}{c|}{12\% Labeled Data}   & \multicolumn{3}{c|}{24\% Labeled Data}             & \multicolumn{3}{c}{49\% Labeled Data}              \\ \cline{2-10} 
\multicolumn{1}{c|}{}     & Task 1          & Task 2          & Task 3                    & Task 1          & Task 2          & Task 3          & Task 1          & Task 2          & Task 3          \\ \hline
Baseline     & 0.5973       & 0.3719       & 0.2784                       & 0.7498          & 0.5413          & 0.3859          & 0.8603          & 0.6066          & 0.5286          \\ \hline
URPC \cite{re:urpc-luo2022semi}        & 0.6649       & 0.4449       & 0.3169         & 0.7486          & 0.5383          & 0.3886          & 0.8361          & 0.6328          & 0.5414          \\
UAMT \cite{re:UAMT}                & \textbf{0.7223}       & 0.2953       & 0.2180        & 0.7760          & 0.5288          & 0.4085          & 0.8811          & 0.6925          & 0.5815          \\
CPS \cite{re:CPS}         & 0.7222       & 0.4570       & 0.3525        & 0.8437          & 0.6562          & 0.4821          & 0.8874          & 0.7774          & 0.5837          \\
CLD \cite{re:CLD-lin2022calibrating}            & 0.7120       & 0.4465       & 0.3056          & 0.8376          & 0.6474          & 0.4325          & 0.8857          & 0.7553          & 0.5999          \\ \hline
\textbf{Co-DA (Ours)}    & 0.7196 & \textbf{0.5023} & \textbf{0.3629}  & \textbf{0.8515} & \textbf{0.6641} & \textbf{0.5040} & \textbf{0.8932} & \textbf{0.7898} & \textbf{0.6475} \\ \hline
\end{tabular}%
\label{table:cadis}
\end{table*}

\begin{figure*}[h!]
\centering
\includegraphics[width=0.6\linewidth]{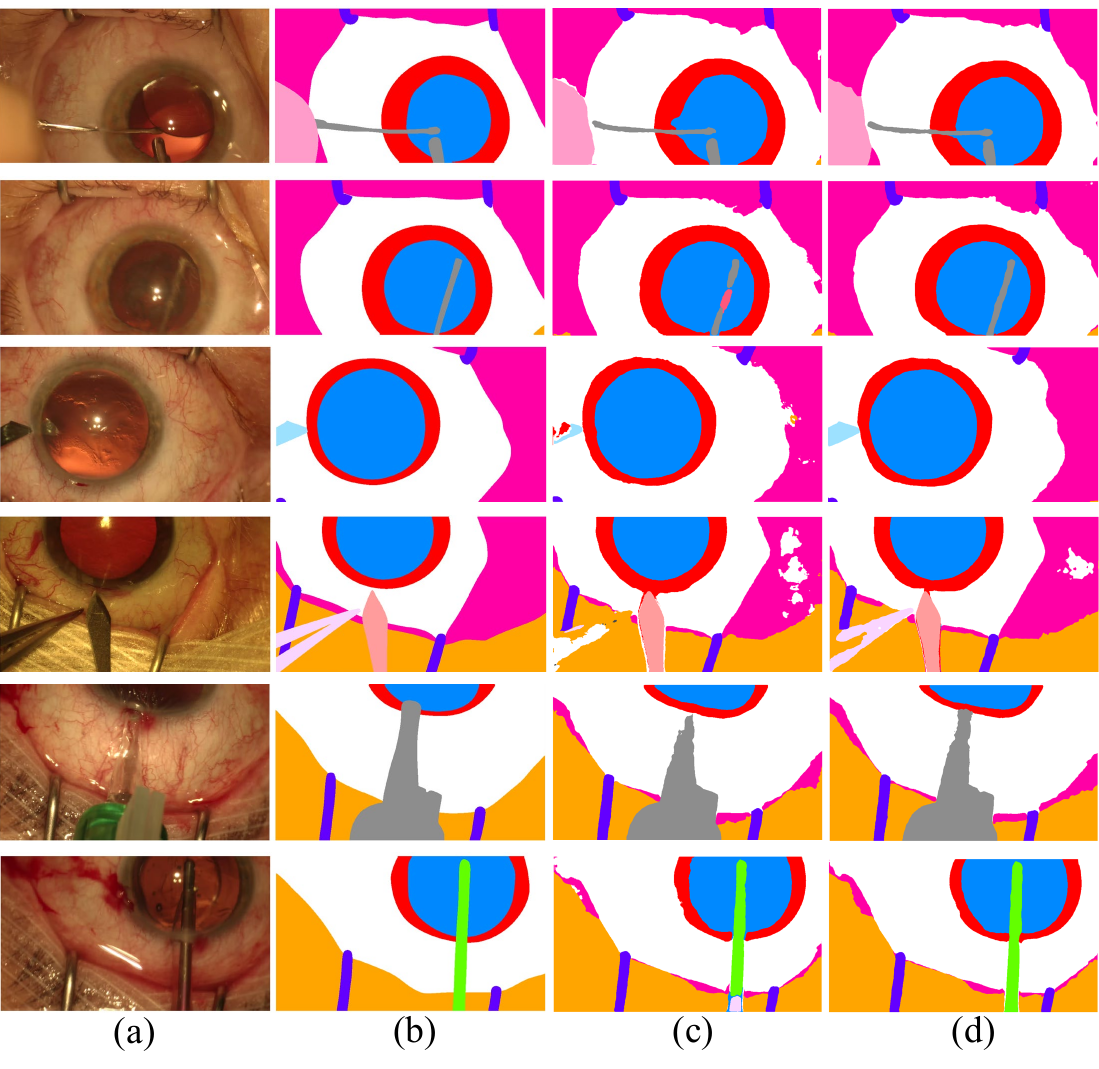}
\caption{Comparison of the proposed Co-DA and another state-of-the-art method, CLD~\cite{re:CLD-lin2022calibrating}, on CaDIS with 49\% labeled data. \textbf{(a)} is the input to the networks, \textbf{(b)} is the ground-truth of the corresponding input, \textbf{(c)} is the output of CLD, and \textbf{(d)} is the output of Co-DA.}
\label{fig:cadis_visual}
\end{figure*}

In Figure~\ref{fig:cadis_visual}, we show the qualitative results of Co-DA. The comparison between Co-DA and CLD \cite{re:CLD-lin2022calibrating} shows that Co-DA could obtain superior results, especially in the instrument regions.

\subsection{Results on Late Gadolinium Enhancement MRI}
\label{sec:res_lge-mri}
\begin{figure}[h!]
\centering
\includegraphics[width=0.5\linewidth]{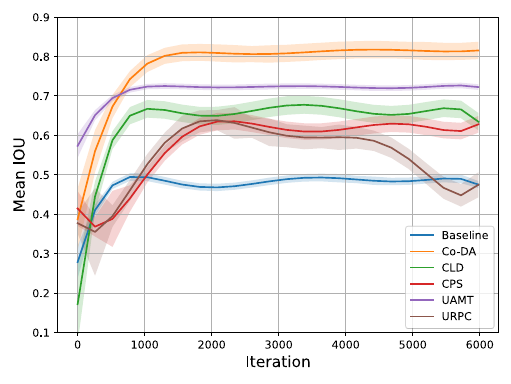}
\caption{Trends in mIoU on the LGE-MRI dataset with 20\% labeled data. It is clear that Co-DA not only provides the best performance but also trains more quickly than most other methods, requiring a smaller number of training iterations. Also, the performance is stable once peaked.}
\label{fig:line}
\end{figure}

Similar to our setup in the CaDIS experiment, the quantitative results of the experiments on {Late Gadolinium Enhancement MRI} (LGE-MRI) with 10\% and 20\% labeled data are shown in Table~\ref{table:la}, where we also present the experimental results for all the competing algorithms. Our proposed Co-DA consistently provides a strong performance among all algorithms. With 10\% labeled data, Co-DA boosts 21.78\% and 26.54\% on Dice and Jaccard over the baseline, and reduces 2.86 voxels and 10.7 voxels on ASD and 95HD, respectively. With 20\% labeled data, Co-DA consistently provides the best performance, improving 25.3\% and 29.48\% on Dice and Jaccard over the baseline, and reduces 5.54 voxels and 16.08 voxels on ASD and 95HD, respectively. We note that our method demonstrates the strongest performance on Dice score and Jaccard under all settings. Furthermore, we present the trends in mIoU with 20\% data and different methods in Figure~\ref{fig:line}. It is evident that the proposed Co-DA trains more quickly than other methods, requiring fewer training iterations than CLD, CPS, URPC, etc. The performance of Co-DA also remains stable once peaked.

\begin{table*}[h!]
\caption{Experimental results on LGE-MRI. Bold numbers represent the best performance.}
\centering
\resizebox{\textwidth}{!}{
\begin{tabular}{l|cccc|cccc}
\hline
\multicolumn{1}{c|}{\multirow{2}{*}{Method}} & \multicolumn{4}{c}{10\% Labeled Data} & \multicolumn{4}{c}{20\% Labeled Data}                                      \\ \cline{2-9} 
\multicolumn{1}{c|}{}     & Dice~(\%)       & Jaccard~(\%)    & ASD~(voxels)    & 95HD~(voxels)      & Dice~(\%)       & Jaccard~(\%)    & ASD~(voxels)    & 95HD~(voxels)    \\ \hline
Baseline   & 62.41          & 46.67          & 10.27          & 36.01        & 62.94          & 49.70          & 9.23          & 30.85          \\ \hline
URPC \cite{re:urpc-luo2022semi}    & 83.67          & 63.57          & 10.27          & 27.67       & 81.30          & 68.90          & 7.14          & 24.43          \\
UAMT \cite{re:UAMT}    & 66.38          & 52.04          & \textbf{6.64}          & \textbf{21.27}       & 86.81          & 76.99          & 4.66          & 17.89          \\
CPS \cite{re:CPS}     & 69.38          & 47.77          & 17.09          & 27.09      & 73.29          & 59.42          & 8.73          & 26.39          \\
CLD \cite{re:CLD-lin2022calibrating}     & 65.22          & 42.26          & 6.68          & 23.79      & 77.03          & 63.85          & 5.27          & 19.66          \\ \hline
\textbf{Co-DA (Ours)}  & \textbf{84.19} & \textbf{73.21} & 7.41 & 25.31    & \textbf{88.24} & \textbf{79.18} & \textbf{3.69} & \textbf{14.77} \\ \hline
\end{tabular}%
}
\label{table:la}
\end{table*}

\begin{figure}[h!]
\centering
\includegraphics[width=0.5\linewidth]{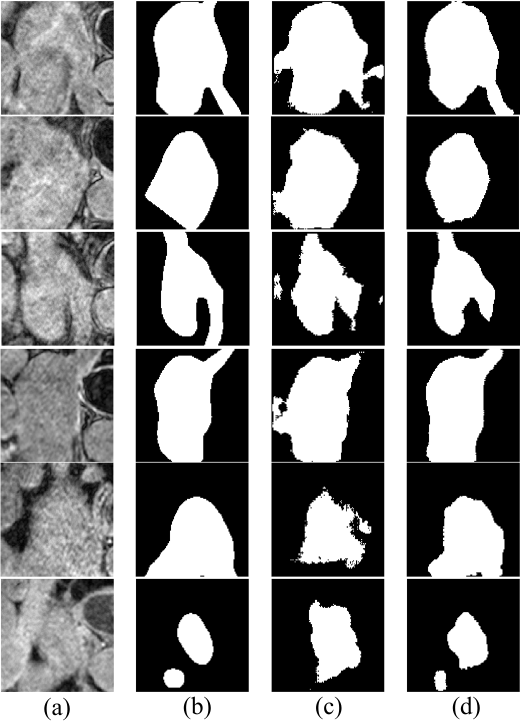}
\caption{Comparison of the proposed Co-DA and another state-of-the-art method, CLD~\cite{re:CLD-lin2022calibrating} on 2D slices of MRI on LGE-MRI with 20\% labeled data, where \textbf{(a)} is the input to the networks, \textbf{(b)} is the corresponding grouth-truth, \textbf{(c)} is the output of CLD and \textbf{(d)} is the output of Co-DA.}
\label{fig:la_visual}
\end{figure}

Figure~\ref{fig:la_visual} shows some examples of the segmentation results on LGE-MRI, using the proposed Co-DA and another state-of-the-art method, CLD~\cite{re:CLD-lin2022calibrating}. CLD tends to have difficulty in accurately segmenting the pathological regions and will partially misjudge the healthy regions, while our segmentation results of the pathological regions are much closer to the ground-truth.  In addition, our method is less likely to generate spurious artefacts commonly found in the results obtained with CLD, showcasing the strong regularization ability of class-wise distribution alignment.

\subsection{Results on ACDC}
\label{sec:res_acdc}
\begin{figure*}[h!]
\centering
\includegraphics[width=0.6\linewidth]{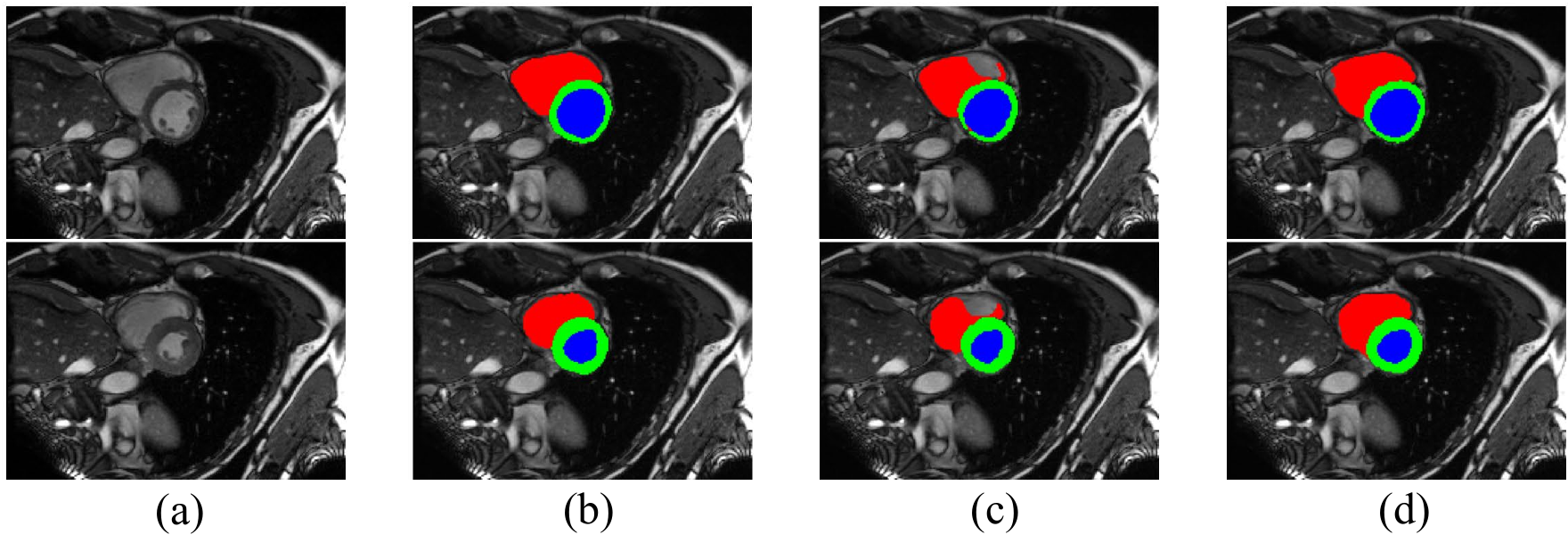}
\caption{Comparison of the proposed Co-DA and another state-of-the-art method, CLD~\cite{re:CLD-lin2022calibrating}, on the ACDC dataset with 10\% labeled data, where \textbf{(a)} is the input to the networks, \textbf{(b)} is the corresponding ground-truth, \textbf{(c)} is the output of CLD and \textbf{(d)} is the ouput of Co-DA.}
\label{fig:acdc_visual}
\end{figure*}

According to Table~\ref{table:acdc}, the performance of our method on the ACDC dataset is better than or comparable to other competing algorithms, which shows that our method is also effective in this segmentation task. With 10\% labeled data, our method is able to outperform the baseline with a 12.35\% margin on Dice and a 6.18 voxels improvement on 95HD. With 20\% labeled data, these performance boosts are 10.3\% and 4.63 voxels, respectively. It is clear that our method provides the best performance in terms of 95HD under both settings. However, our method, along with other methods for semi-supervised medical image segmentation, {displays a slightly poorer performance} than CPS on Dice score, indicating the excellent performance of CPS and showing that distribution alignment for pseudo-labels does not work well enough on the ACDC dataset, which we will continue to explore in future work. Although our method does not provide the best Dice score, the performance in terms of 95HD is better than all other algorithms, demonstrating that Co-DA is able to provide high quality segmentation with accurate boundary delineation. We present example segmentation results of CLD and our proposed Co-DA in Figure~\ref{fig:acdc_visual}.

\begin{table*}[h!]
\caption{Experimental results on ACDC. Bold numbers represent the best performance.}
\centering
\setlength{\tabcolsep}{4mm}
\begin{tabular}{l|cc|cc}
\hline
\multicolumn{1}{c|}{\multirow{2}{*}{Method}} & \multicolumn{2}{c|}{10$\%$ Labeled Data}                           & \multicolumn{2}{c}{20$\%$ Labeled Data}                           \\ \cline{2-5} 
\multicolumn{1}{c|}{}                        & Dice~(\%) & 95HD~(voxels) & Dice(\%) & 95HD~(voxels) \\ \hline
Baseline                                     & 70.59                    & 12.11                     & 77.43                    & 9.57                     \\ \hline
URPC\cite{re:urpc-luo2022semi}                                         & 77.59                    & 6.46                      & 86.07                    & 5.07                     \\
UAMT\cite{re:UAMT}                                         & 72.47                    & 15.49                     & 82.68                    & 6.05                     \\
CPS\cite{re:CPS}                                          & \textbf{83.96}                    & 8.75                      & \textbf{87.81}                    & 5.95                     \\
CLD\cite{re:CLD-lin2022calibrating}                                          & 83.04                    & 6.13                      & 86.58                    & 5.81                     \\ \hline
\textbf{Co-DA(Ours)}                                 & 82.94           & \textbf{5.93}             & 87.73           & \textbf{4.94}            \\ \hline
\end{tabular}%
\label{table:acdc}
\end{table*}

\subsection{Ablation Studies}
\label{sec:abl}

In this section, we perform additional experiments to verify the strong learning capacity Co-DA as compared to its fully supervised variant. In addition, we isolate the different components of our method to ensure that they all contribute to the final performance. Finally, we look into the efficacy of the dynamic threshold for the over-expectation cross-entropy loss and compare examples of pseudo-labels generated with and without Co-DA.

\begin{figure}[H]
\centering
\includegraphics[width=0.4\linewidth]{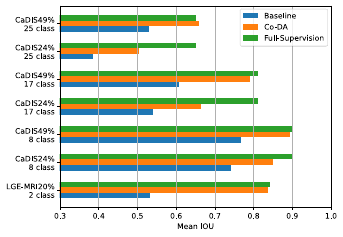}
\caption{Comparing the baseline, Co-DA, and the fully supervised method on the CaDIS dataset in terms of mIoU. In some cases, Co-DA performs comparably to the fully supervised method, demonstrating its strong learning capacity.}
\label{fig:full}
\end{figure}

Firstly, we compare Co-DA with the fully supervised settings on both CaDIS and LGE-MRI and present the results in Figure~\ref{fig:full}. In many cases, the performance of Co-DA is very close to the fully supervised setting, demonstrating the outstanding efficacy of class-wise distribution alignment. Specifically, with 49\% labeled data, Co-DA provides only slightly lower mIoU on CaDIS Tasks 1 and 2 as compared to the fully supervised setting. It even marginally outperforms the fully supervised case on Task 3, which means the refined pseudo-labels have a superior quality.

Secondly, to ensure the complementary effect of proposed components in our method, i.e., the co-distribution alignment and the over-expectation cross-entropy loss, we conduct a set of ablation experiments on CaDIS, and the results are presented in Table \ref{table:abl}. The third row demonstrates that the over-expectation cross-entropy loss reduces the noise in pseudo-labels by dynamically varying the threshold to filter out low-confidence pixels, achieving a significant improvement of 0.0488 on Task 3 for mean IoU over the baseline. The fourth row shows that the proposed Co-DA can bring a gain of 0.1303 on Task 3, while there is a slight decrease in performance on Task 2. Empirically, we conjecture the reason is that Task 1 and Task 3 have more severe class imbalance than Task 2 \cite{re:cadis-bouget2017vision}. Considering the overall stability of the proposed method, we apply the two newly designed modules to co-training simultaneously, obtaining improvements of 0.0058, 0.0124 and 0.0748 on the three tasks over the baseline, respectively.

\begin{table}[h!]
\caption{Results from our ablation study on CaDIS. Performance shown in mean IoU. O-E denotes the proposed over-expectation cross entropy loss.}
\centering
\setlength{\tabcolsep}{1.25mm}
\begin{tabular}{cccc|ccc}
\hline
\multirow{2}{*}{Co-training} & \multirow{2}{*}{DA} & \multirow{2}{*}{O-E} & \multirow{2}{*}{Co-DA} & \multicolumn{3}{c}{49\% Labeled Data} \\ \cline{5-7} 
  &   &   &   & \multicolumn{1}{c}{Task1}     & \multicolumn{1}{c}{Task2}     & Task3     \\ \hline
\checkmark &   &   &   & \multicolumn{1}{c}{0.8874} & \multicolumn{1}{c}{0.7774} & 0.5837 \\
\checkmark &  \checkmark &   &   & \multicolumn{1}{c}{0.8893} & \multicolumn{1}{c}{0.7718} & 0.6438 \\
\checkmark &   &  \checkmark &   & \multicolumn{1}{c}{0.8903} & \multicolumn{1}{c}{0.7824} & 0.6325 \\
\checkmark &   &   & \checkmark & \multicolumn{1}{c}{0.8907} & \multicolumn{1}{c}{0.7681} & \textbf{0.7140} \\
\checkmark &   & \checkmark & \checkmark & \multicolumn{1}{c}{\textbf{0.8932}} & \multicolumn{1}{c}{\textbf{0.7898}} & 0.6585 \\ \hline
\end{tabular}%
\label{table:abl}
\end{table}

\begin{figure}[H]
\centering
\includegraphics[width=0.5\linewidth]{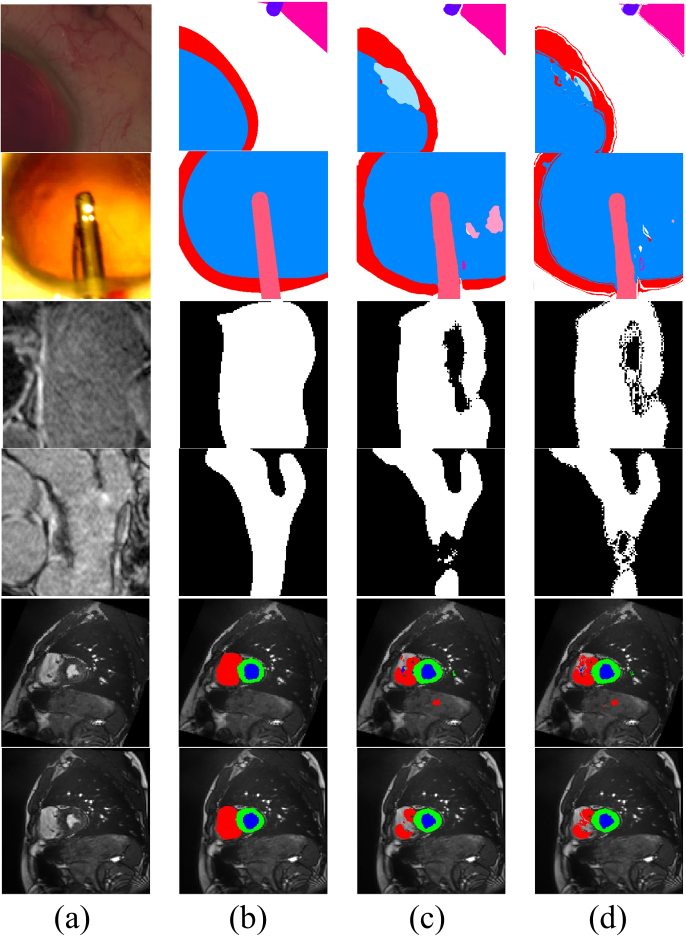}
\caption{
Visualization of pseudo-labels generated by Co-DA. \textbf{(a)} is the input image, \textbf{(b)} is the Ground-Truth, \textbf{(c)} is pseudo-labels generated by the network without Co-DA, and \textbf{(d)} is the pseudo-labels refined by Co-DA.
}
\label{fig:pseudo}
\end{figure}

We also visualize examples of the pseudo-labels generated by the networks with and without Co-DA, and the results are shown in Figure \ref{fig:pseudo}. Clearly, Co-DA is able to improve the quality of pseudo-labels, resulting in an overall accuracy improvement and a reduction of the negative impact from incorrect pseudo-labels.

\begin{figure}[h!]
\centering
\includegraphics[width=0.4\linewidth]{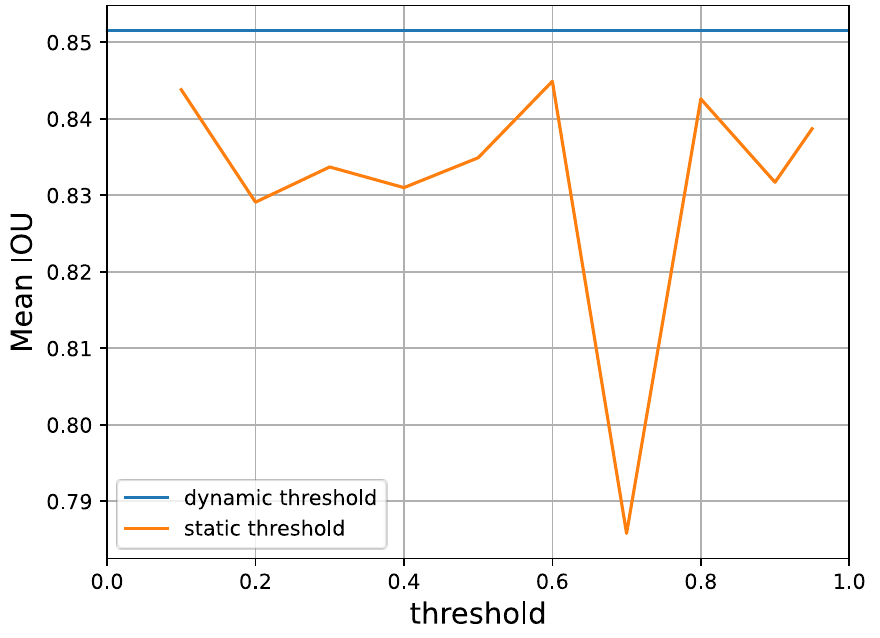}
\caption{
A performance comparison of static thresholds versus the dynamic threshold that our method adopts according to Equation~\ref{condition_threshold} on Task 1 of the CaDIS dataset with 24\% labeled data. It is clear that the dynamic threshold provides superior mean IoU.
}
\label{fig:abl}
\end{figure}

Finally, according to Figure~\ref{fig:abl}, different predefined thresholds for the over-expectation loss will also lead to different levels of performance. Yet, a dynamic threshold as provided by Equation~\eqref{condition_threshold} is consistently superior, as it is able to automatically select a suitable gating value for retaining reliable pseudo-labels.

\section{Conclusions}
\label{sec:conclusion}

In this paper, we propose a novel Co-Distribution Alignment (Co-DA) approach to partially labeled medical image segmentation tasks with imbalanced class distributions. Our key idea involves a class-dependent alignment between labeled distributions and unlabeled marginal distributions that is suitable for dense prediction tasks. Specifically, Co-DA simplifies the estimation of labeled distributions, extends the original DA to class-wise distribution alignment with cross-supervision and adopts adaptive temperature scaling for labeled distributions to avoid highly imbalanced estimations. In addition, we propose an over-expectation cross-entropy loss to reduce noises in pseudo-labels. Extensive experiments, including abation studies, on three publicly available datasets demonstrate the consistently superior learning capacity of our approach.

\vspace{5mm}
\noindent \textbf{Funding.} This research was funded by National Natural Science Foundation of China (61972187, 61703195), Fujian Provincial Natural Science Foundation (2022J011112, 2020J02024, 2020J01828), Research Project of Fashu Foundation (MFK23001) and The Open Program of The Key Laboratory of Cognitive Computing and Intelligent Information Processing of Fujian Education Institutions, Wuyi University (KLCCIIP2020202).

\bibliographystyle{ieee}
\bibliography{references}

\end{document}